\documentclass[conference]{IEEEtran}
\IEEEoverridecommandlockouts
\usepackage{cite}
\usepackage{amsmath,amssymb,amsfonts}
\usepackage{algorithmic}
\usepackage{graphicx}
\usepackage{textcomp}
\usepackage{xcolor}
\usepackage{placeins}
\def\BibTeX{{\rm B\kern-.05em{\sc i\kern-.025em b}\kern-.08em
    T\kern-.1667em\lower.7ex\hbox{E}\kern-.125emX}}

\usepackage{booktabs} 
\usepackage{threeparttable}
\usepackage{listings}

\definecolor{codegreen}{rgb}{0,0.6,0}
\definecolor{codegray}{rgb}{0.5,0.5,0.5}
\definecolor{codepurple}{rgb}{0.58,0,0.82}
\definecolor{backcolour}{rgb}{0.95,0.95,0.92}

\lstdefinestyle{mystyle}{
    backgroundcolor=\color{backcolour},   
    commentstyle=\color{codegreen},
    keywordstyle=\color{magenta},
    numberstyle=\tiny\color{codegray},
    stringstyle=\color{codepurple},
    basicstyle=\ttfamily\footnotesize,
    breakatwhitespace=false,         
    breaklines=true,                 
    captionpos=b,                    
    keepspaces=true,                 
    numbers=left,                    
    numbersep=5pt,                  
    showspaces=false,                
    showstringspaces=false,
    showtabs=false,                  
    tabsize=2
}
\usepackage{tikz}
\usetikzlibrary{matrix, positioning, backgrounds, fit, calc}
\usepackage{amsmath}  
\usepackage{xcolor}
\usepackage{array}
\usepackage{booktabs} 
\usepackage{acronym}
\usepackage{easyReview}
\usepackage{hyperref}

\newcommand{\vx}[1]{\vec{#1}}
\newcommand{\mx}[1]{\mathbf{#1}}

\colorlet{colqset}{orange!30}
\colorlet{colkvset}{green!30}
\colorlet{colaset}{blue!20}
\colorlet{colpset}{purple!25}
\colorlet{notset}{white!30}

\definecolor{forestgreen}{rgb}{0.13, 0.55, 0.13}
\colorlet{colkvlearn}{forestgreen!80}
\colorlet{colalearn}{blue!50}
\colorlet{colplearn}{purple!55}

\usepackage{pythonhighlight}

\begin{document}
\acrodef{AC}[AC]{Arrenhius \& Current}
\acrodef{AER}[AER]{Address Event Representation}
\acrodef{AEX}[AEX]{AER EXtension board}
\acrodef{AMDA}[AMDA]{``AER Motherboard with D/A converters''}
\acrodef{API}[API]{Application Programming Interface}
\acrodef{BM}[BM]{Boltzmann Machine}
\acrodef{CAVIAR}[CAVIAR]{Convolution AER Vision Architecture for Real-Time}
\acrodef{CCN}[CCN]{Cooperative and Competitive Network}
\acrodef{CD}[CD]{Contrastive Divergence}
\acrodef{CMOS}[CMOS]{Complementary Metal--Oxide--Semiconductor}
\acrodef{COTS}[COTS]{Commercial Off-The-Shelf}
\acrodef{CPU}[CPU]{Central Processing Unit}
\acrodef{CV}[CV]{Coefficient of Variation}
\acrodef{CV}[CV]{Coefficient of Variation}
\acrodef{DAC}[DAC]{Digital--to--Analog}
\acrodef{DBN}[DBN]{Deep Belief Network}
\acrodef{DFA}[DFA]{Deterministic Finite Automaton}
\acrodef{DFA}[DFA]{Deterministic Finite Automaton}
\acrodef{divmod3}[DIVMOD3]{divisibility of a number by 3}
\acrodef{DPE}[DPE]{Dynamic Parameter Estimation}
\acrodef{DPI}[DPI]{Differential-Pair Integrator}
\acrodef{DSP}[DSP]{Digital Signal Processor}
\acrodef{DVS}[DVS]{Dynamic Vision Sensor}
\acrodef{EDVAC}[EDVAC]{Electronic Discrete Variable Automatic Computer}
\acrodef{EIF}[EI\&F]{Exponential Integrate \& Fire}
\acrodef{EIN}[EIN]{Excitatory--Inhibitory Network}
\acrodef{EPSC}[EPSC]{Excitatory Post-Synaptic Current}
\acrodef{EPSP}[EPSP]{Excitatory Post--Synaptic Potential}
\acrodef{FPGA}[FPGA]{Field Programmable Gate Array}
\acrodef{FSM}[FSM]{Finite State Machine}
\acrodef{GPU}[GPU]{Graphical Processing Unit}
\acrodef{HAL}[HAL]{Hardware Abstraction Layer}
\acrodef{HH}[H\&H]{Hodgkin \& Huxley}
\acrodef{HMM}[HMM]{Hidden Markov Model}
\acrodef{HW}[HW]{Hardware}
\acrodef{hWTA}[hWTA]{Hard Winner--Take--All}
\acrodef{IF2DWTA}[IF2DWTA]{Integrate \& Fire 2--Dimensional WTA}
\acrodef{IF}[I\&F]{Integrate \& Fire}
\acrodef{IFSLWTA}[IFSLWTA]{Integrate \& Fire Stop Learning WTA}
\acrodef{INCF}[INCF]{International Neuroinformatics Coordinating Facility}
\acrodef{INI}[INI]{Institute of Neuroinformatics}
\acrodef{IO}[IO]{Input-Output}
\acrodef{IPSC}[IPSC]{Inhibitory Post-Synaptic Current}
\acrodef{ISI}[ISI]{Inter--Spike Interval}
\acrodef{JFLAP}[JFLAP]{Java - Formal Languages and Automata Package}
\acrodef{LIF}[LI\&F]{Linear Integrate \& Fire}
\acrodef{LSM}[LSM]{Liquid State Machine}
\acrodef{LTD}[LTD]{Long-Term Depression}
\acrodef{LTI}[LTI]{Linear Time-Invariant}
\acrodef{LTP}[LTP]{Long-Term Potentiation}
\acrodef{LTU}[LTU]{Linear Threshold Unit}
\acrodef{MCMC}{Markov Chain Monte Carlo}
\acrodef{NHML}[NHML]{Neuromorphic Hardware Mark-up Language}
\acrodef{NMDA}[NMDA]{NMDA}
\acrodef{NME}[NE]{Neuromorphic Engineering}
\acrodef{PCB}[PCB]{Printed Circuit Board}
\acrodef{PRC}[PRC]{Phase Response Curve}
\acrodef{PSC}[PSC]{Post-Synaptic Current}
\acrodef{PSP}[PSP]{Post--Synaptic Potential}
\acrodef{RI}[KL]{Kullback-Leibler}
\acrodef{RRAM}[RRAM]{Resistive Random-Access Memory}
\acrodef{RBM}[RBM]{Restricted Boltzmann Machine}
\acrodef{ROC}[ROC]{Receiver Operator Characteristic}
\acrodef{RNN}[RNN]{Recurrent Neural Network}
\acrodef{SAC}[SAC]{Selective Attention Chip}
\acrodef{SCD}[SCD]{Spike-Based Contrastive Divergence}
\acrodef{SCX}[SCX]{Silicon CorteX}
\acrodef{STDP}[STDP]{Spike Time Dependent Plasticity}
\acrodef{SW}[SW]{Software}
\acrodef{sWTA}[SWTA]{Soft Winner--Take--All}
\acrodef{VHDL}[VHDL]{VHSIC Hardware Description Language}
\acrodef{VLSI}[VLSI]{Very  Large  Scale  Integration}
\acrodef{WTA}[WTA]{Winner--Take--All}
\acrodef{XML}[XML]{eXtensible Mark-up Language}
\acrodef{MLP}[MLP]{multilayer perceptron}
\acrodef{ReLU}[ReLU]{rectified linear unit}
\acrodef{VMM}[VMM]{Vector Matrix Multiplication}
\acrodef{CNN}[CNN]{Convolutional Neural Network}

\title{On-Chip Learning via Transformer In-Context Learning} 

\author{\IEEEauthorblockN{Jan Finkbeiner\IEEEauthorrefmark{1}\IEEEauthorrefmark{2}, Emre Neftci\IEEEauthorrefmark{1}\IEEEauthorrefmark{2}}
\IEEEauthorblockA{\IEEEauthorrefmark{1}Fakult\"at für Elektrotechnik und Informationstechnik, RWTH Aachen, Aachen, 52074, Germany}
\IEEEauthorblockA{\IEEEauthorrefmark{2}Peter Gr\"unberg Institut, Forschungszentrum J\"ulich GmbH, J\"ulich, 52425, Germany}
\IEEEauthorblockA{\{j.finkbeiner, e.neftci\}@fz-juelich.de}
\and
}


\maketitle

\begin{abstract}
Autoregressive decoder-only transformers have become key components for scalable sequence processing and generation models.
However, the transformer's self-attention mechanism requires transferring prior token projections from the main memory at each time step (token), thus severely limiting their performance on conventional processors.
Self-attention can be viewed as a dynamic feed-forward layer, whose matrix is input sequence-dependent similarly to the result of local synaptic plasticity. 
Using this insight, we present a neuromorphic decoder-only transformer model that utilizes an on-chip plasticity processor to compute self-attention. 
Interestingly, the training of transformers enables them to ``learn'' the input context during inference.
We demonstrate this \emph{in-context learning} ability of transformers on the Loihi 2 processor by solving a few-shot classification problem. 
With this we emphasize the importance of pretrained models especially their ability to find simple, local, backpropagation free, learning rules enabling on-chip learning and adaptation in a hardware friendly manner.
\end{abstract}


\section{Introduction}
\acp{RNN} and in particular transformers have become essential components for a wide variety of scalable sequence processing models \cite{Goodfellow_etal16_deeplear,Vaswani_etal17_atteall,Gu_Dao23_mambline,Yang_etal23_gateline,Beck_etal24_xlstexte}.
Transformers and their \ac{RNN}-based variants have drawn significant attention thanks to their ability to be massively scaled \cite{Kaplan_etal20_scallaws, hoffmann2022trainingcomputeoptimallargelanguage} and their ability to be used as foundation models~\cite{Bommasani_etal21_opporisk}. 
Their emergent ability to learn new input-output functions during inference, also called in-context learning, is believed to be at the heart of their performance capabilities \cite{Oswald_etal23_uncomesa}.

During inference, the self-attention mechanism of transformers requires loading prior token caches from memory, scaling poorly with the sequence length and leading to very low arithmetic intensity (i.e. few compute per memory operations) \cite{Kim_etal23_fullstac}. 
Thus, there is a growing interest in designing dedicated in and near-memory processors to accelerate transformer models \cite{Leroux_etal24_analinme,Wu_etal24_pimgpt,Sridharan_etal23_xforinme,Laguna_etal21_inmecomp}.

Transformer models consist almost exclusively of operations that are easily realizable on neuromorphic substrates in a local fashion. Specifically, they do not require weight sharing or non-linearities except for the normalization and softmax operation and activation functions.

Chips designed for in- and near-memory processing enable the local storage of token caches, specifically the key and value representations. This allows for immediate reuse of those representations as weight matrices during subsequent token processing steps, eliminating the need to transfer them to other parts of the chip or retrieve them from external memory. Additionally, autoregressive inference avoids the instantiation of a full attention mask, which typically scales quadratically with sequence length, thereby significantly reducing peak memory requirements. These advantages underscore the potential of in- and near-memory processing for autoregressive transformer inference, while also introducing the need for dynamic weight-memory adjustment mechanisms.

Furthermore, several studies have demonstrated activation sparsity that naturally arises from softmax and \ac{ReLU} activation functions~\cite{li2023lazyneuronphenomenonemergence, mirzadeh2023relustrikesback, alizadeh2024llmflash, liu2024trainingfreeactivationsparsitylarge} which can be leveraged for more efficient processing in event-based architectures.

Transformers' emergent ability to learn new input-output functions during inference, also called in-context learning, is believed to be at the heart of their performance.
Analyzing a linearized self-attention mechanism \cite{Akyurek_etal22_whatlear,Oswald_etal23_tranlear} showed that they have the capability of performing gradient descent on least squares problems \cite{Garg_etal22_whatcan}. 
This line of work further demonstrated that attention weight updates are indeed aligned with such gradients, and can be generalized to more than one layer \cite{Oswald_etal23_uncomesa}.

If inference in transformers is a learning process, then the training of transformers is a meta-learning process that can solve few-shot learning problems \cite{Brown_etal20_langmode}.
Interpreting transformer training and inference as a meta-learning setup reveals the simplicity of the learned inference rule, which reduces to straightforward row-wise updates of weight matrices via linear projections of the current representation. Crucially, this rule operates without the need for backpropagation, loss-projection layers, or other external mechanisms, while still allowing weight updates across all layers, even in deep networks.

In summary, the transformer’s inference rule acts as a fully local plasticity mechanism with minimal hardware requirements.
Using this understanding, we implemented autoregressive transformers on Loihi 2~\cite{Davies_etal18_loihneur}, a digital neuromorphic near-memory computing chip with event-based communication and processing, by leveraging its local learning engine. We demonstrate the transformer’s meta- and few-shot learning capabilities, along with its potential for on-chip learning, by solving a standard few-shot learning task based on the Omniglot dataset~\cite{Lake_etal15_humaconc}.

\subsection{Related Work}
Prior work \cite{Stewart_etal24_emulbrai} demonstrated few-shot learning of visual patterns on the Intel Loihi 1 chip using Model Agnostic Meta Learning (MAML) on spike-based convolutional networks. 
MAML computes the gradients of a local plasticity rule on a training task set and optimizes the network for one-shot learning. 
\cite{Hajizada_etal24_contlear} demonstrate a related few-shot online continual learning approach that learns new categories by adding prototypes.
In contrast to the above work, our approach uses a transformer network and in principle being generalizable to any sequence learning problem, whereas prior work on the Intel Loihi chip were limited to classification problems. 
Beyond digital architectures, several works~\cite{Karunaratne_etal21_robuhigh, Mao_etal22_expevali, Li_etal21_sapi64kb} explored the applicability of emerging memories for Associative Memories (AMs) and Content Addressable Memories (CAMs) to enable few-shot learning and adaptation via Memory augmented neural networks (MANNs). Contrary to MANNs, the transformer architecture has been demonstrated to be predictably scalable~\cite{Kaplan_etal20_scallaws, hoffmann2022trainingcomputeoptimallargelanguage} achieving greater performance at larger scale, while serving the same promise of a few-shot, in-context adaptable architecture.

\section{Methods}
%
\subsection{Autoregressive Decoder-Only Transformer Inference}
\begin{figure}[t]
\centering
\includegraphics[width=.8\linewidth]{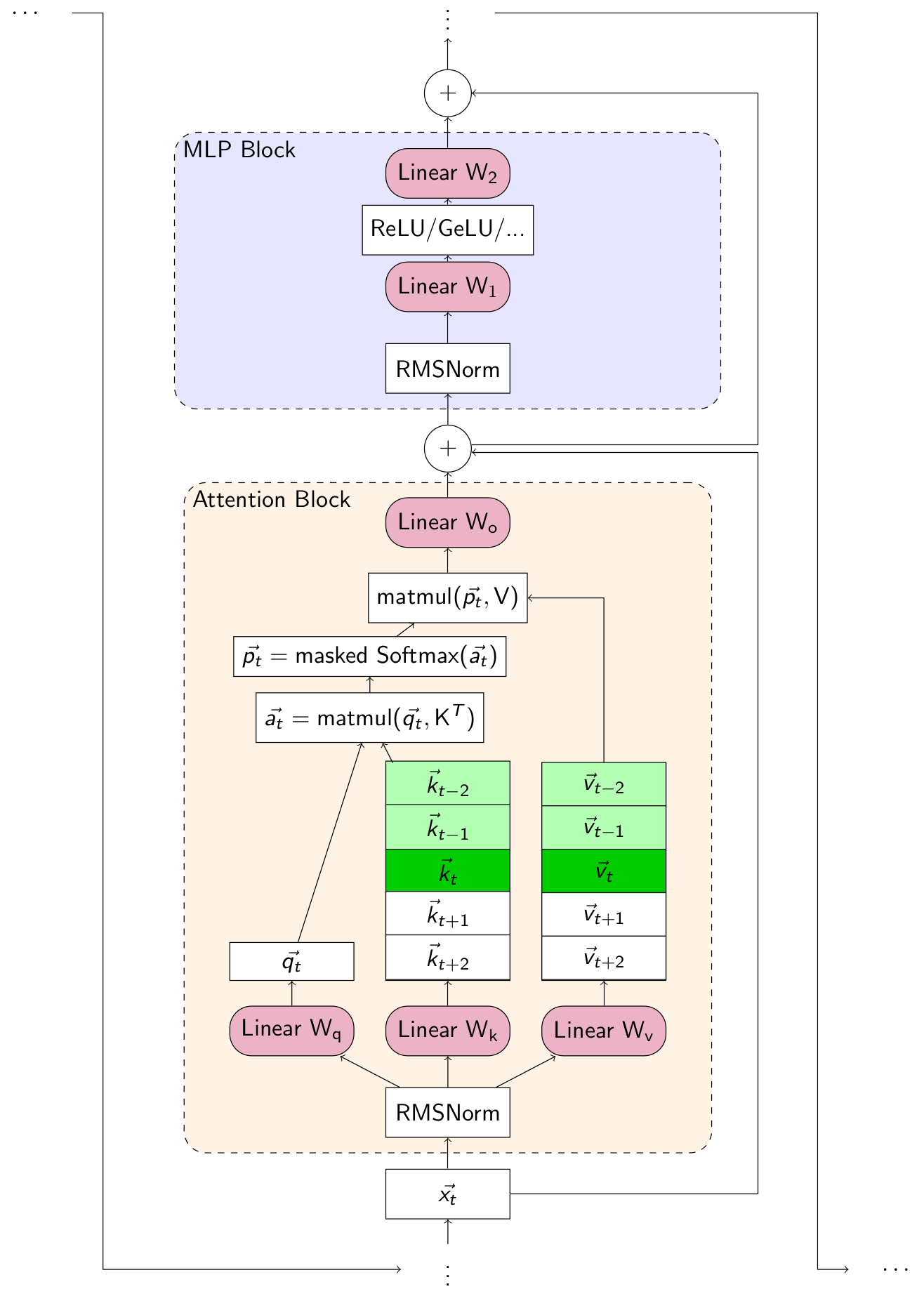}
\caption{\label{fig:transformersetup} Illustration of an Autoregressive Decoder-Only Transformer layer.}
\end{figure}
\begin{figure}[t]
\centering
\includegraphics[width=.72\linewidth]{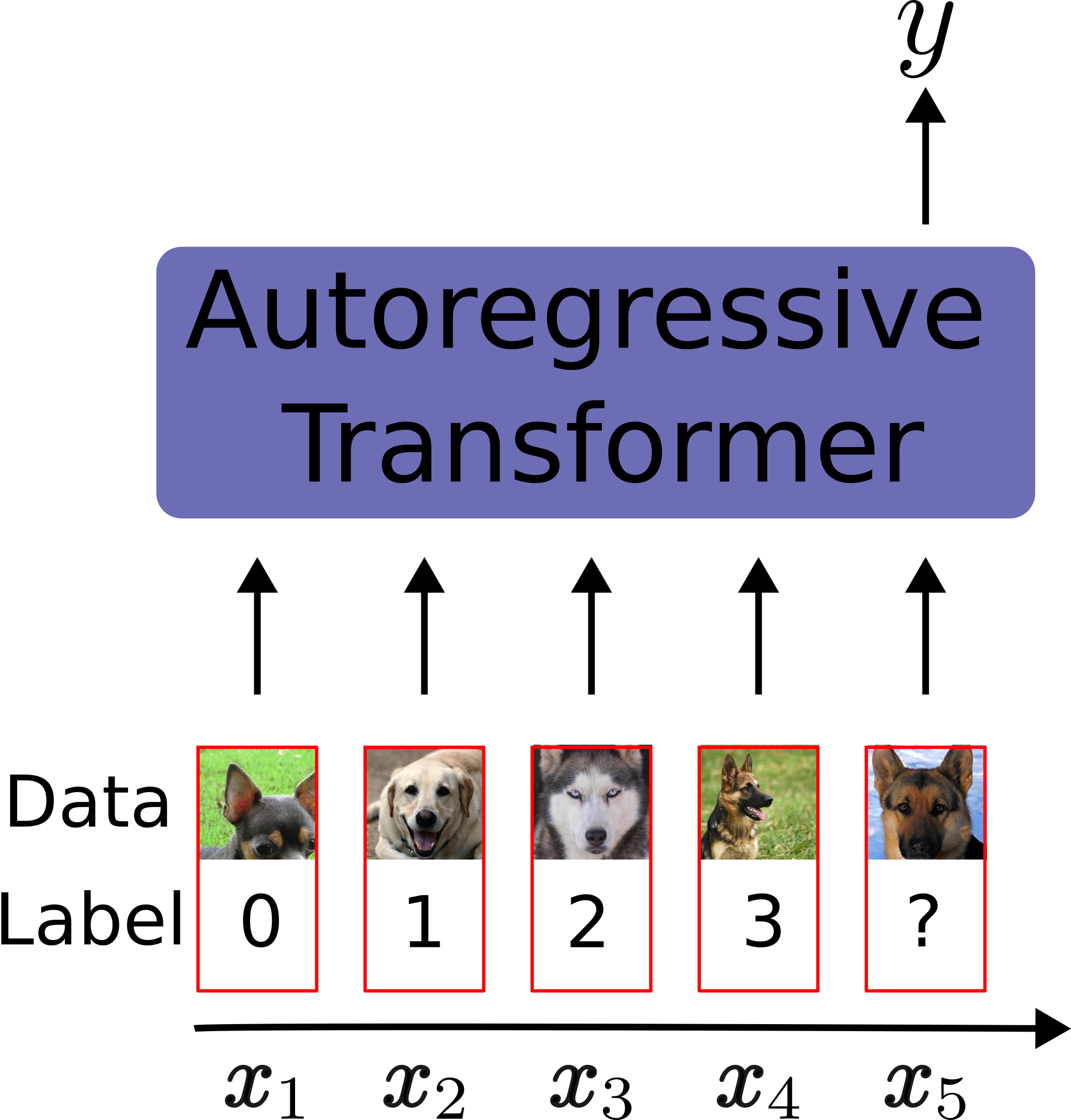}
\caption{\label{fig:fewshotlearningsetup} Illustration of the problem setup for the $N$-way $K$-shot few-shot learning experiment. Following prior work \cite{Vinyals_etal16_matcnetw}, a support set of $N$ image samples is associated with $K$ arbitrary labels. In our experiments, we used the Omniglot dataset~\cite{Lake_etal15_humaconc} instead of dog breeds.}
\end{figure}
\begin{figure*}[ht]
\centering
\includegraphics[width=0.4\linewidth]{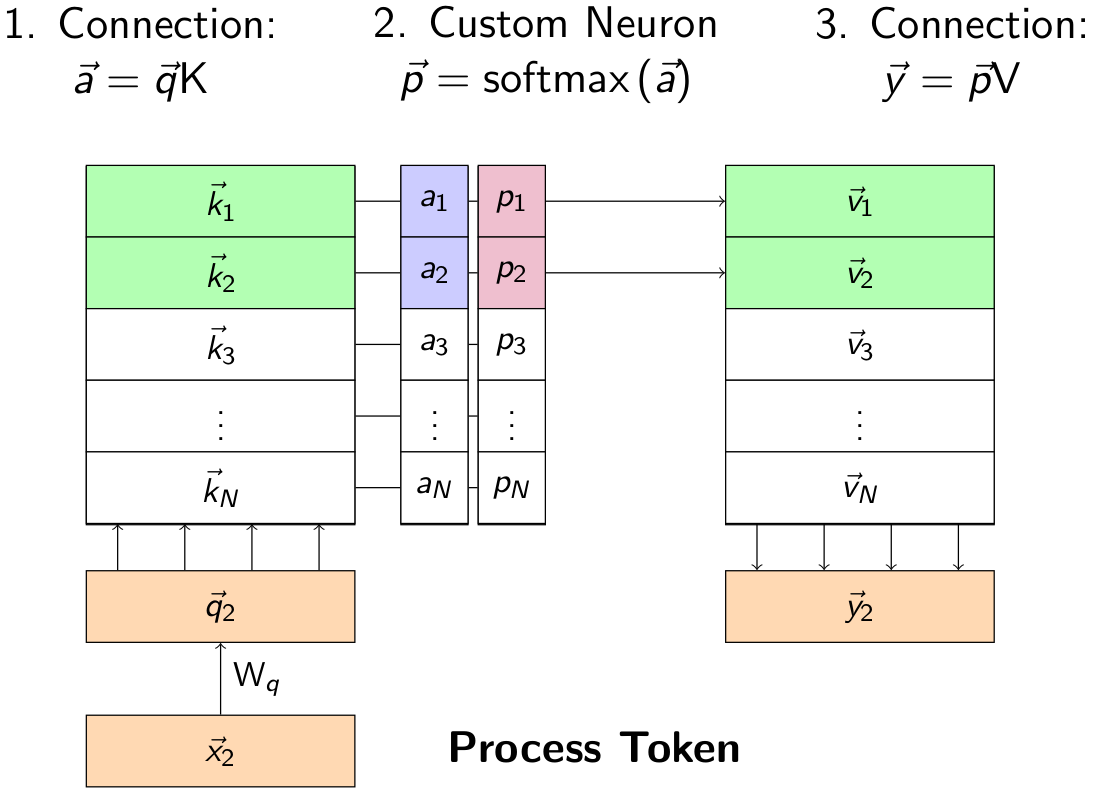}
\includegraphics[width=0.44\linewidth]{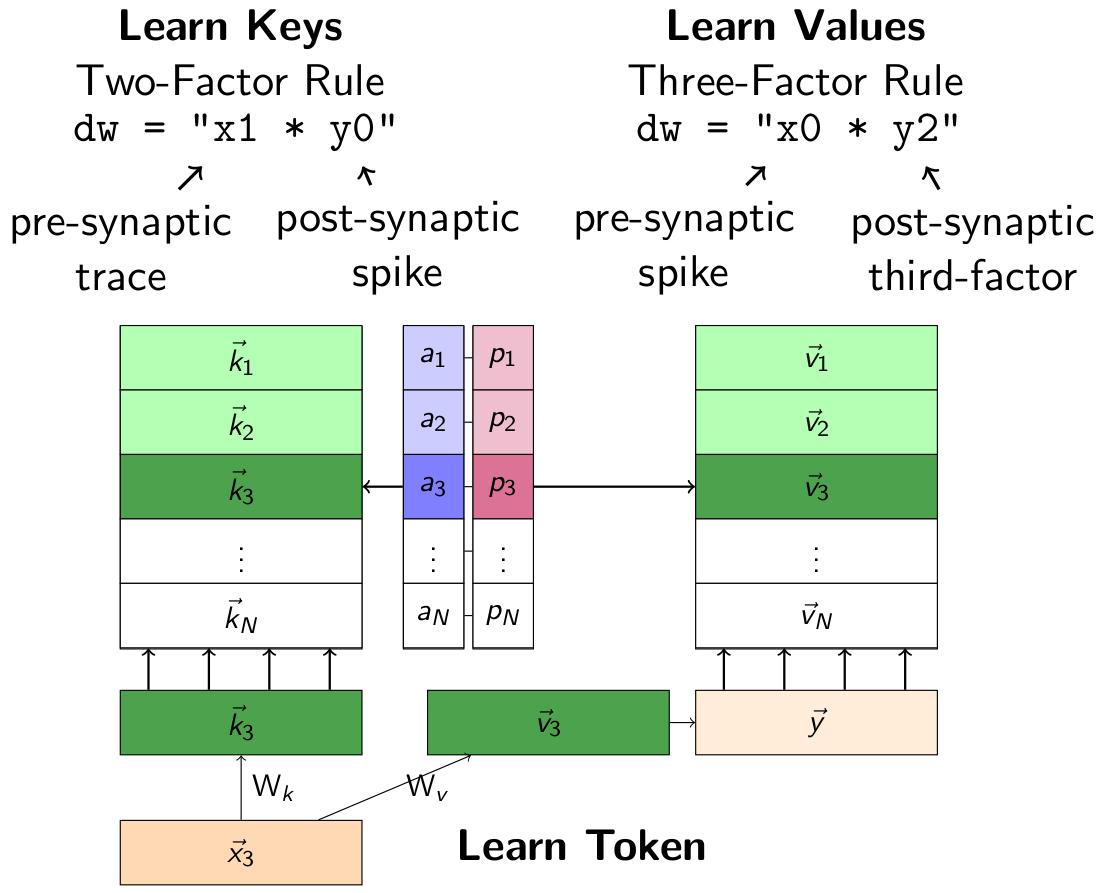}
\caption{Illustration demonstrating the interpretation of the self-attention mechanism as local plasticity rule.}
\label{fig:loihilearning}
\end{figure*}
In this work, we focus on decoder-only transformer architectures with a causal attention mask (Fig.~\ref{fig:transformersetup}). Each transformer layer, connected via residual connections, consists of a self-attention block and an \ac{MLP} block. The operations of a single-head attention block ($H=1$) with embedding dimension $D$ at time $t \leq T$ are described in Equations \ref{eq:rmsmha}-\ref{eq:outmha}:
\begin{align}
    \vx{x}_t^\prime &= \text{RMSNorm} \left( \vx{x}_t \right) \; &\in \mathbb{R}^{D}, \label{eq:rmsmha} \\
    \left[ \vx{q}_t , \vx{k}_t, \vx{v}_t \right] &= \left[ \mx{W}_q , \mx{W}_k \mx{W}_v \right] \; \vx{x}_t^\prime \; &\in \mathbb{R}^{3D}, \\
    \vx{a}_t &= \vx{q}_t \, \left[ \vx{k}_0, \vx{k}_1, \dots, \vx{k}_t \right]^T =  \vx{q}_t \, \mx{K}_{[0:t]}  \; &\in \mathbb{R}^{t}, \label{eq:vmmkcache}\\
    \vx{p}_t &= \text{softmax} \left( \vx{a}_t \right)  \; &\in \mathbb{R}^{t}, \\
    \vx{y}_t &= \vx{p}_t \, \left[ \vx{v}_0, \vx{v}_1, \dots, \vx{v}_t \right] = \vx{p}_t \, \mx{V}_{[0:t]} \; &\in \mathbb{R}^{D}, \label{eq:vmmvcache} \\
    \vx{z}_t &= \mx{W}_o \; \vx{y}_t + \vx{b}_o \; &\in \mathbb{R}^{D} , \label{eq:outmha}
\end{align}
where the key and value matrices, $\mx{K}_{[0:t]}$ and $\mx{V}_{[0:t]}$, from the so called KV-cache. Equations \ref{eq:rmsmlp}-\ref{eq:outmlp} describe the typical two-layer \ac{MLP} block with a \ac{ReLU} as activation function: 
\begin{align}
    \vx{x}_t^\prime &= \text{RMSNorm} \left( \vx{x}_t \right) \quad &\in \mathbb{R}^{D}, \label{eq:rmsmlp} \\
    \vx{y}_t &= \text{ReLU} \left( \mx{W}_1 \; \vx{x}_t^\prime \right) \quad &\in \mathbb{R}^{4D}, \\
    \vx{z}_t &= \mx{W}_2 \; \vx{y}_t + \vx{b}_2 \quad &\in \mathbb{R}^{D}.  \label{eq:outmlp}
\end{align}
Transformer variants with both LayerNorm and RMSNorm exist. We chose to use the RMSNorm for it's simplicity.
For training, the input sequence is processed in parallel with a causal attention mask, leading to an attention matrix $\mx{A} \in \mathbb{R}^{T \times T \times H}$ that scales quadratically in sequence length. During inference, the sequence can be processed in a token-by-token autoregressive manner without ever instantiating the full attention matrix. Thereby, autoregressive transformer inference resembles the inference mode of a recurrent neural network with a matrix state, the KV-cache. 

We implemented the transformer layer as described by Equations \ref{eq:rmsmha}-\ref{eq:outmlp}, including support for multiple attention heads, on Intel's Loihi 2 chip~\cite{Davies_etal18_loihneur}. The KV-cache and corresponding dynamic \acp{VMM} are implemented as described in Sec. \ref{sec:selfattn_as_locplacticity}. The normalization, softmax and ReLU operation are implemented with custom neurons that handle division and exponential operations written in Lava-Loihi's microcode.
%
\subsection{Self-Attention as Local Plasticity Rules} \label{sec:selfattn_as_locplacticity}
In this section, we reformulate the dynamic self-attention operation, specifically the construction of the KV-cache, as a local plasticity rule for weight matrices. This approach reveals the similarity between plasticity rules and the self-attention mechanism, highlighting the local nature of the emerging meta-learned learning rule.
Figure \ref{fig:loihilearning} (left) shows that for queries $\vx{q}_t$ and attention weights $\vx{p}^t$, the KV caches, $\mx{K}_{[0:t]}$ and $\mx{V}_{[0:t]}$, act as standard weight matrices for \acp{VMM} (Eq. \ref{eq:vmmkcache}, \ref{eq:vmmvcache}).
According the the input-output relation described in Equations \ref{eq:vmmkcache} and \ref{eq:vmmvcache} we define pre- and post-synaptic terms. 
For the Keys-VMM (Eq. \ref{eq:vmmkcache}), $\vx{q}_t$ and $\vx{k}_t$ represent the pre-synaptic partner's activations, $\vx{a}^t$ the post-synaptic partner's state. During the plasticity process, this translates to $\vx{k}_t$ representing a pre-synaptic trace \texttt{x1}, whereas the corresponding neurons $\vx{a}_i$ can send a post-synaptic spike \texttt{y0} to trigger the learning process of it's synapses. 
Similarily, for the Values-VMM (Eq. \ref{eq:vmmvcache}) we can define $\vx{p}$ as pre-synaptic neurons and $\vx{y}$ as post-synaptic neuron activations. Here, the values to be learned, $\vx{v}$, are sent to the post-synaptic neurons and enter the plasticity rule as a secondary, external factor \texttt{y2}. The corresponding neuron $\vx{p}_i$ sends a pre-synaptic spike \texttt{x0} to trigger the weight update.

Summarizing, we formulate the Key-learning rule as a two factor learning rule with a pre-synaptic trace variable \texttt{x1} and a post-synaptic spike trigger \texttt{y0}, and the Value-learning rule as a three factor learning rule with pre-synaptic spike trigger \texttt{x0} and post-synaptic external trace variable \texttt{y2}, allowing for synapse local weight updates.

To demonstrate the applicability and usefulness of the presented local plasticity formulation we implement self-attention on Loihi 2~\cite{Davies_etal18_loihneur} using it's learning rule engine with aforementioned local plasticity formulation. Loihi 2 is a digital near-memory computing chip with event based communication and processing capabilities. 
Network architectures can be created by connecting ``Connection'' and ``Neuron'' processes with each other. 
It contains a programmable learning rule engine, which supports local plasticity rules based on pre-and post-synaptic spikes and trace values that can be combine in a sum-of-product form. 
All projection layers are implemented as \texttt{Dense} connection process, that operates as matrix for \acp{VMM}. Listing \ref{lst:loihilearning} shows the setup of learning rules and connection processes related to the dynamic part of the self-attention operation. For the plastic key and value matrices we  instantiate \texttt{LearningDense} processes. The keys-connection is equipped with a \texttt{GradedSpikeCfg.OVERWRITE} configuration in order to write the the pre-synaptic $\vx{k}_t$ events into the pre-synaptic traces \texttt{x1}. The keys- and values-learning rule, \texttt{lr\_keys} and \texttt{lr\_values}, implement the operations described above, adjusted for Loihi's constraint of unsigned traces. The addition of the terms ``\texttt{- y0 * w}'' and ``\texttt{- x0 * y1 * w}'' allows for the implementation of sliding window attention~\cite{iz20longformer}, where previous keys and values are overwritten with a first-in first-out mechanism.
%
\begin{figure}[t] 
    \centering
\begin{lstlisting}[language={python}, caption={Implementation of the learning connection processes for the KV-caches of self-attention in the programming framework Lava for Loihi 2.}, label={lst:loihilearning}]
# keys learning rule
lr_keys = Loihi2FLearningRule(
    dw="2 * y0 * (x1 - 64) - y0 * w", ...)

# keys cache connection process
conn_keys = LearningDense(
    weights=inital_keys,  graded_spike_cfg=GradedSpikeCfg.OVERWRITE,
    learning_rule=lr_keys, ...)

# values learning rule
lr_values = Loihi3FLearningRule(
    dw="x0 * y2 - x0 * y3 - x0 * y1 * w", ...)

# values cache connection process
conn_values = LearningDense(
    weights=inital_values,
    learning_rule=lr_values, ...)
\end{lstlisting}
\end{figure}
%
\subsection{Meta-Learning with Transformers Setup}
We demonstrate our model on a N-Way K-shot classification problem, where the model is trained on $K$ samples, or shots, of $N$ categories, out of all of the available samples and classes in a dataset. The goal of this few-shot learning method is to train a model that can classify new categories using $K$ samples.
The experimental design of the N-way K-Shot learning problem follows \cite{Vinyals_etal16_matcnetw}, where a support set consisting of $K$ shots of $N$ categories is presented, followed by a query image and constant ``query'' token. 
A similar setup was used in \cite{Chan_etal22_datadist}, except here that images and labels are concatenated into the same token as depicted in Figure \ref{fig:fewshotlearningsetup}.

We use a simple setup with a single hidden layer and normalization to encode the input (a concatenation of the flattened image and label) into an embedded token. This token is then passed through a multi-layer decoder-only transformer. A linear projection generates the output of dimension $N$, and class probabilities are obtained by applying a softmax operation on the final representation during training.

\section{Results}
Results of the few-shot learning experiment are shown in Table \ref{tbl:omniglot}. 
We compare our results with MAML~\cite{Finn_etal17_modemeta}, an established bi-level optimization method for few-shot learning, Matching Nets~\cite{Vinyals_etal16_matcnetw} and MANN~\cite{santoro16mann}. 
We use the standard Omniglot dataset which contains contains 1623 categories of 28x28 pixel handwritten characters, and 20 samples for each category written by Amazon’s Mechanical Turk participants~\cite{Lake_etal15_humaconc}. 
The Lava and Loihi results were obtained on a limited set of 32-256 task samples and are therefore still effected by sampling noise. 
Overall the results demonstrate the competitiveness of such a simple transformer based few-shot approach compared to gradient based approaches, specifically MAML~\cite{Finn_etal17_modemeta}. 
The results from Lava and Loihi were obtained using a limited set of 32-256 task samples, which introduces some sampling noise. It's important to note that the results were not derived from a clearly defined test set, making direct comparisons between experiments and methods challenging. Overall, the findings demonstrate that this simple transformer-based few-shot approach is competitive with gradient-based methods, particularly MAML~\cite{Finn_etal17_modemeta}.

We primarily analyze a single-head, 4-layer transformer network with an embedding dimension of $D=128$, while also including an experiment with a slightly larger model to explore scalability for improved task performance. We test four model variants: a floating-point baseline, a Loihi-compatible quantization, a fixed-point Lava implementation, and a Loihi deployment, where 2 consecutive layers are executed on Loihi and two on a CPU. The Loihi-compatible quantized networks perform almost on par with the floating-point baseline. Although the fixed-point Lava simulations and the Loihi-deployed models show slight accuracy degradation, particularly on the more complex 20-way task, they still demonstrate competitive performance or even outperform non-convolutional network approaches. The accuracy drop does not indicate a fundamental disadvantage but rather underscores the need for careful deployment, potentially through improved quantization-aware pretraining methods or scaling. Notably, several studies have shown that inference at very low precision is feasible for large-scale networks and large language models (LLMs)~\cite{xiao2024smoothquantaccurateefficientposttraining, lin2024awqactivationawareweightquantization, ma2024era1bitllmslarge, zhu2024scalablematmulfreelanguagemodeling}.



%
\begin{table}[t]
    \caption{Few-shot classification on held-out Omniglot characters for architectures without (top) and with (bottom) convolutions. Results show a floating point baseline (Float), Loihi compatible quantization (Quant), fixed point lava implementation (Lava) and the Loihi 2 implementation (Loihi).}
    \label{tbl:omniglot}
    \begin{threeparttable}
    \centering
    \begin{tabular}{ l c c c c }
    \toprule
    \multicolumn{1}{ c}{} & \multicolumn{2}{ c }{5-way Accuracy} & \multicolumn{2}{ c }{20-way Accuracy}\\
    { Omniglot~\cite{Lake_etal15_humaconc}}  &  1-shot &  5-shot &  1-shot &  5-shot\\
    \midrule
    MANN, no conv~\cite{santoro16mann} & $82.8\%$ & $94.9\%$ &-- & -- \\
    MAML, no conv~\cite{Finn_etal17_modemeta} & $ 89.7 \%$ & $ 97.5 \%$ & -- & -- \\
    Transformer-tiny\tnote{a}  \; (Loihi) & $ 86.1 \%$ & $ 96.3 \%$ & $ 60.9 \% $ & $ 83.4 \% $ \\
    Transformer-tiny\tnote{a}  \; (Lava)  & $ 90.2 \%$ & $ 94.1 \%$ & $ 68.4 \% $ & $ 86.6 \% $ \\
    Transformer-tiny\tnote{a}  \; (Quant) & $ 90.3 \%$ & $ 97.1 \%$ & $ 76.7 \% $ & $ 88.4 \% $ \\
    Transformer-tiny\tnote{a}  \; (Float) & $ 93.1 \%$ & $ 97.3 \%$ & $ 77.7 \% $ & $ 89.2 \% $ \\
    Transformer-small\tnote{b} \; (Float) & $ 95.8 \%$ & $ 98.3 \%$ & $ 82.8 \% $ & $ 94.2 \% $ \\
    \midrule
    Matching Nets~\cite{Vinyals_etal16_matcnetw}  & $98.1 \%$ & $98.9 \%$  & $93.8 \%$ & $98.5 \%$ \\
    MAML~\cite{Finn_etal17_modemeta}  & $98.7 \%$ & $99.9 \%$ & $95.8 \%$ & $98.9 \%$ \\
    \bottomrule
    \end{tabular}
    \begin{tablenotes}\footnotesize
    \item[a] 4 transformer layers, embed dim 128, 1 head
    \item[b] 6 transformer layers, embed dim 256, 8 heads
    \end{tablenotes}
    \end{threeparttable}
\end{table}
\section{Discussion}
Large-scale pretrained models, which exhibit in-context learning behaviors, are effectively meta-trained to use a simple learning rule that can be formulated as a local plasticity process and efficiently implemented on neuromorphic platforms. We leverage Loihi 2’s learning engine to achieve this kind of inference-time learning, showcasing the duality between inference and learning.
Importantly, the transformer architecture has been shown to be a salable architecture with predictable performance improvements as a function of parameter count and training data~\cite{Kaplan_etal20_scallaws, hoffmann2022trainingcomputeoptimallargelanguage}. Hardware development can greatly benefit from such a model's capabilities as well as their modularity and scalability.
In conclusion, we advocate for the benefits of a symbiotic relationship between scalable neural network architectures, particularly the autoregressive transformer architecture, and neuromorphic hardware as a step towards lifelong on-device learning.
\section*{Acknowledgment}
This work was sponsored by the Federal Ministry of Education, Germany BMBF under grants no. 16ME0398K, 16ME0399, 01IS22094E; and Neurosys as part of the initiative "Cluster4Future" funded by the Federal Ministery of Education and Research BMBF (03ZU1106CB). The project was funded by the Federal Ministry of Education and Research (BMBF) under grant no. 01IS22094E WEST-AI. We thank the support of Intel Labs in providing the Loihi2 system and helping with fundamental questions regarding the Loihi architecture and it's programmability. Lastly, we thank Alpha Renner for moral support and insightful tips.

 \section*{Copyright}
© 2024 IEEE. Personal use of this material is permitted. Permission from IEEE must be obtained for all other uses, in any current or future media, including reprinting/republishing this material for advertising or promotional purposes, creating new collective works, for resale or redistribution to servers or lists, or reuse of any copyrighted component of this work in other works.

\FloatBarrier
{\footnotesize

\bibliographystyle{plain}
}
\end{document}